\documentclass[letterpaper]{article} 
\usepackage{aaai2026}  
\usepackage{times}  
\usepackage{helvet}  
\usepackage{courier}  
\usepackage[hyphens]{url}  
\usepackage{graphicx} 
\usepackage{natbib}  
\usepackage{caption} 
\usepackage{algorithm}
\usepackage{algorithmic}

\usepackage{newfloat}
\usepackage{listings}
\usepackage{amsmath}
\usepackage{multirow}
\usepackage{subfigure}
\usepackage{amssymb}
\usepackage{tikz}

\DeclareCaptionStyle{ruled}{labelfont=normalfont,labelsep=colon,strut=off} 
\floatstyle{ruled}
\newfloat{listing}{tb}{lst}{}
\floatname{listing}{Listing}
\title{ReflexDiffusion: Reflection-Enhanced Trajectory Planning for High-lateral-acceleration Scenarios in Autonomous Driving}
\author{
    Xuemei Yao\textsuperscript{\rm 1}, Xiao Yang\textsuperscript{\rm 2,3}\thanks{Corresponding authors}, 
Jianbin Sun\textsuperscript{\rm 1*}, Liuwei Xie\textsuperscript{\rm 2,3}, Xuebin Shao\textsuperscript{\rm 4}, \\ Xiyu Fang\textsuperscript{\rm 4}, Hang Su\textsuperscript{\rm 2,3}, Kewei Yang\textsuperscript{\rm 1}\\}
\affiliations{
    \textsuperscript{\rm 1}College of Systems Engineering, National University of Defense Technology\\
    \textsuperscript{\rm 2} Department of Computer Science and Technology, Institute for AI, BNRist Center, Tsinghua University \\ 
    \textsuperscript{\rm 3} Tsinghua-Bosch Joint ML Center \\
    \textsuperscript{\rm 4} CATARC Intelligent Technology \\

    {\{yaoxuemei2019,sunjianbin,kayyang27\}@nudt.edu.cn, yangxiao19@tsinghua.org.cn, lxiear@outlook.com, \{shaoxuebin,fangxiyu\}@catarc.ac.cn, suhangss@tsinghua.edu.cn}
}

\usepackage{bibentry}
\usepackage{url}

\begin{document}

\maketitle

\begin{abstract}
Generating safe and reliable trajectories for autonomous vehicles in {long-tail scenarios} remains a significant challenge, particularly for \textbf{high-lateral-acceleration maneuvers} such as sharp turns, which represent critical safety situations. Existing trajectory planners exhibit systematic failures in these scenarios due to data imbalance. This results in insufficient modelling of vehicle dynamics, road geometry, and environmental constraints in high-risk situations, leading to suboptimal or unsafe trajectory prediction when vehicles operate near their physical limits.
In this paper, we introduce \textbf{ReflexDiffusion}, a novel inference-stage framework that enhances diffusion-based trajectory planners through reflective adjustment. Our method introduces a gradient-based adjustment mechanism during the iterative denoising process: after each standard trajectory update, we compute the gradient between the conditional and unconditional noise predictions to explicitly amplify critical conditioning signals, including road curvature and lateral vehicle dynamics. This amplification enforces strict adherence to physical constraints, particularly improving stability during high-lateral-acceleration maneuvers where precise vehicle-road interaction is paramount. Evaluated on the nuPlan Test14-hard benchmark, ReflexDiffusion achieves a 14.1\% improvement in driving score for high-lateral-acceleration scenarios over the state-of-the-art (SOTA) methods. This demonstrates that inference-time trajectory optimization can effectively compensate for training data sparsity by dynamically reinforcing safety-critical constraints near handling limits. The framework's architecture-agnostic design enables direct deployment to existing diffusion-based planners, offering a practical solution for improving autonomous vehicle safety in challenging driving conditions. 
\end{abstract}


\begin{links}
\link{Code}{https://github.com/Luminous2028/ReflexDiffusion.git}
\end{links}

\section{Introduction}
Safe autonomous driving fundamentally depends on generating reliable trajectories in long-tail scenarios. High-lateral-acceleration maneuvers presenting a critical paradox: they exhibit the highest accident risk yet receive the lowest representation in training data\cite{choudhari_risk_2021,antonios_vehicles_2023,rafiei_effect_2024,kota_analyzing_2025,sun_development_2025}. While recent advances have focused on generating synthetic edge cases to augment training datasets\cite{tang_bev-tsr_2025,xu_diffscene_2025,peng_safety-critical_2025}, a crucial gap remains in fully leveraging existing data to improve planning quality during inference. Current trajectory planning approaches fall into two categories: rule-based methods employ manually designed constraints\cite{dauner_parting_nodate,treiber_congested_nodate} but fail to adapt to novel scenarios; learning-based approaches that utilize imitation\cite{cheng_pluto_2024,cheng_rethinking_2024} or reinforcement learning\cite{liu_multi-task_2022,liu_mtd-gpt_2023} struggle to capture the multi-modal nature of real-world driving, often resulting in suboptimal planning policies.

Diffusion models have emerged as a promising solution for trajectory planning. For instance, the Diffusion Planner \cite{zheng_diffusion-based_2025} jointly models agent dynamics, lane topology, and navigation information using a transformer architecture. However, these approaches exhibit critical limitations in \textbf{high-lateral-acceleration scenarios}, where they often generate unsafe trajectories due to curvature-speed decoupling. Furthermore, they typically rely on carefully designed classifier guidance functions. These functions must be continuously differentiable while satisfying multiple constraints-such as passenger comfort and collision avoidance, significantly complicating deployment.

The field of large language models offers a compelling paradigm through iterative reflection mechanisms. In this approach, generate-evaluate-refine cycles dramatically enhance output quality for high-stakes tasks that require complex constraint satisfaction \cite{ji_towards_2023,shinn_reflexion_2023,madaan_self-refine_2023,gou_critic_2024}.
This paradigm holds significant promise for high-lateral-acceleration trajectory planning, where vehicles must operate near their physical limits while maintaining strict adherence to centripetal force constraints. Although reflection-based approaches have shown success across numerous domains \cite{liu_chain_2023, bai_zigzag_2024}, their application to trajectory planning remains unexplored. This gap presents an untapped opportunity to enhance safety when vehicles operate near their handling limits.

In this paper, we present ReflexDiffusion, a physics-aware reflection framework that pioneers a new approach for trajectory generation in autonomous driving. ReflexDiffusion operates through a dual-phase inference process: standard denoising generates an initial trajectory, which is then refined by a novel reflection phase that injects physics-aware gradients to explicitly enforce curvature-speed-acceleration coupling. This is achieved through conditional gradient ascent that amplifies critical physical constraints, iteratively refining the trajectory until convergence. Specifically, at each denoising step, we compute the gradient difference between conditional and unconditional noise predictions. This operation reveals how physical constraints influence trajectory generation. By strategically amplifying and injecting these gradients, we compel the trajectory to strictly adhere to centripetal force limitations, preventing dangerous speed-curvature violations that plague existing methods. Crucially, this approach eliminates the need for hand-crafted classifier guidance while maintaining full compatibility with existing diffusion-based planners, requiring no structural modifications. Thus, ReflexDiffusion fundamentally transforms the planning paradigm: rather than treating physical constraints as external penalties, it embeds them into the generative process through gradient-based reflection, enabling robust safety-critical planning tasks.

We demonstrate that ReflexDiffusion achieves state-of-the-art (SOTA) performance on high-lateral-acceleration scenarios in the nuPlan Test14-hard \cite{cheng_rethinking_2024} and Test14-random \cite{cheng_rethinking_2024} benchmarks. The method excels in safety-critical edge cases such as U-turns, where curvature-speed coupling is paramount. Our main contributions are as follows:
    \begin{itemize}
    \item \textbf{Pioneering physics-aware reflection for trajectory generation.} We introduce the first adaptation of the reflection mechanism to autonomous driving trajectory planning. Our framework leverages gradient-based adjustment to perform iterative trajectory correction during diffusion sampling, thereby addressing critical safety gaps in real-time planning.
    
    \item \textbf{Breakthrough performance in high-lateral-acceleration scenarios.} By explicitly amplifying curvature-speed coupling through physics-aware reflection, ReflexDiffusion achieves a \textbf{14.1\% improvement} in driving score on high-lateral-acceleration scenarios within the Test14-hard benchmark, setting a new state-of-the-art.
    
    \item \textbf{Architecture-agnostic generalization.} As a plug-and-play module, the framework boosts the driving score of the Diffusion-es planner by \textbf{22.5\%} on Test14-hard when deployed during inference, demonstrating compelling universality and transferability without any architectural modifications.
\end{itemize}

\section{Related Works}
\textbf{Autonomous vehicles trajectory generation.}
Trajectory generation for autonomous vehicles is primarily addressed by rule-based and learning-based methods. Rule-based approaches use explicit logical rules derived from traffic and safety knowledge \cite{treiber_congested_nodate,dauner_parting_nodate}, but lack generalization in complex scenarios. Learning-based methods, such as imitation learning \cite{cheng_pluto_2024} and reinforcement learning \cite{shalev-shwartz_safe_2016,liu_mtd-gpt_2023}, learn policies from data but often produce suboptimal, unimodal trajectories. Diffusion models address this by capturing multi-modal behavior, however, they frequently depend on post-processing or guidance functions \cite{yang_diffusion-es_nodate,zheng_diffusion-based_2025}, which compromises their flexibility and interpretability. Our framework overcomes these limitations by eliminating the need for explicit guidance or rule-based refinement, instead enabling direct and flexible trajectory adaptation through a novel diffusion reflection mechanism.

\textbf{Challenges in long-tail scenarios.}
Long-tail scenarios involving high lateral acceleration (e.g., emergency curve negotiation) pose significant challenges to existing planners. These challenges stem from data scarcity, the oversimplified constraints of rule-based methods, and the static risk adaptation in imitation learning \cite{becker_model-_2023,yoshioka_path_2025}. Existing approaches often fail to incorporate key physical principles, such as curvature-speed-acceleration coupling ($a_y = \kappa v^2$), resulting in unstable vehicle behavior near handling limits. Our method bridges this gap by leveraging inference-time reflection to dynamically reinforce $\kappa$-$v$ relationship and adjust safety margins using real-time lateral acceleration feedback.

\textbf{Reflection mechanisms.}
Reflection mechanisms iteratively refine outputs through generate-evaluate-adjust cycles, improving correctness and safety in domains such as large language models and computer vision \cite{liu_chain_2023,madaan_self-refine_2023, bai_zigzag_2024}. However, they remain underexplored in trajectory planning, especially for high-dynamic scenarios. Existing adaptations in this domain are either computationally prohibitive for real-time use or fail to guarantee adherence to physical constraints \cite{yang_diffusion-es_nodate,zheng_diffusion-based_2025}. Our work introduces a real-time reflection mechanism that employs physics-compliant gradient adjustment and safety-gated triggers to proactively correct trajectories upon detecting curvature-speed misalignment.

\textbf{Generalization of trajectory generation methods.}
Generalization under distribution shift remains a persistent challenge. Rule-based methods typically require manual re-engineering per scenario, while data-driven generative models, such as GANs \cite{jiang_continuous_2023} and the Diffusion Planner \cite{zheng_diffusion-based_2025}, often lack inference-time adaptability. In contrast, our reflection framework is architecture-agnostic. This design enables zero-shot transfer to unseen planners and delivers improved performance on nuPlan, without any structural modifications to the base model.

\section{Problem Formulation}
\textbf{Task Definition.} High-lateral-acceleration scenarios represent a critical subset of long-tail driving situations characterized by extreme vehicle dynamics near handling limits. Formally, a driving scenario $\mathcal{S}$ is classified as high-lateral-acceleration \cite{karnchanachari_towards_nodate} if:

\begin{equation}
\exists t \in [t_0, t_0 + \Delta t],\quad |a_y(t)| \geq a_{\text{th}}, \quad \Delta t \geq t_{\text{min}}
\label{eq:acc_threshold}
\end{equation}
where $a_y(t)$ is the instantaneous lateral acceleration (m/s²); $a_{\text{th}} = 4.0$ m/s² is the acceleration threshold; $\Delta t \geq 0.5$ s is the minimum duration (nuPlan benchmark).

Conventional planners exhibit critical failures in these scenarios due to curvature-speed decoupling. The curvature $\kappa$ of the planned trajectory and the vehicle speed $v$ fail to satisfy the centripetal force constraint $a_y = \kappa v^2$. A violation occurs when $|\kappa v^2 - a_y| > 4.0\ \text{m/s}^2, \kappa=1/R$, resulting in understeer or oversteer, where $\kappa$ represents planned trajectory curvature and $R$ represents turning radius.

Our objective is to mitigate this decoupling problem by explicitly reinforcing the physical coupling between curvature, velocity, and acceleration through the reflection mechanism, using the same joint modelling approach as in Diffusion Planner \cite{zheng_diffusion-based_2025}, which enables interaction-aware planning. \\
   \textbf{Inputs.} The inputs comprise (1) Ego state: Current position $(x,y)$, heading $(\cos\theta,\sin\theta)$; (2) Agent history: Neighbor agents' states over $L=21$ past timesteps, and their attributes include type (car, bicycle and pedestrian), size and velocity. (3) Scene context; (4) Vectorized HD Map: Lane polylines with traffic lights and speed limits; (5) Static obstacles: Construction zones, road barriers; (6) Route navigation: Sequence of route lanes. \\
\textbf{Outputs.} The output formation is a joint trajectory matrix including ego trajectory as well as $M$ neighbors trajectories in the next 8 seconds at 10Hz, i.e., $\textbf{x}^{(0)} \in \mathbb{R}^{(M+1) \times \tau \times 4}$:
\[
\textbf{x}^{(0)} = 
\begin{bmatrix}
x_{\text{ego}}^{1} &  \cdots & x_{\text{ego}}^{\tau} \\
x_{\text{neigh}_1}^{1} & \cdots & x_{\text{neigh}_1}^{\tau} \\
\vdots & \ddots & \vdots \\
x_{\text{neigh}_M}^{1} & \cdots & x_{\text{neigh}_M}^{\tau}
\end{bmatrix}, 
\;
x_{\text{agent}}^k = 
\begin{bmatrix} 
x \\ y \\ \cos\theta \\ \sin\theta
\end{bmatrix}
\]
   \section{Methodology}
\begin{figure*}[htbp]
        \centering
        \includegraphics[width=0.9\textwidth]{
            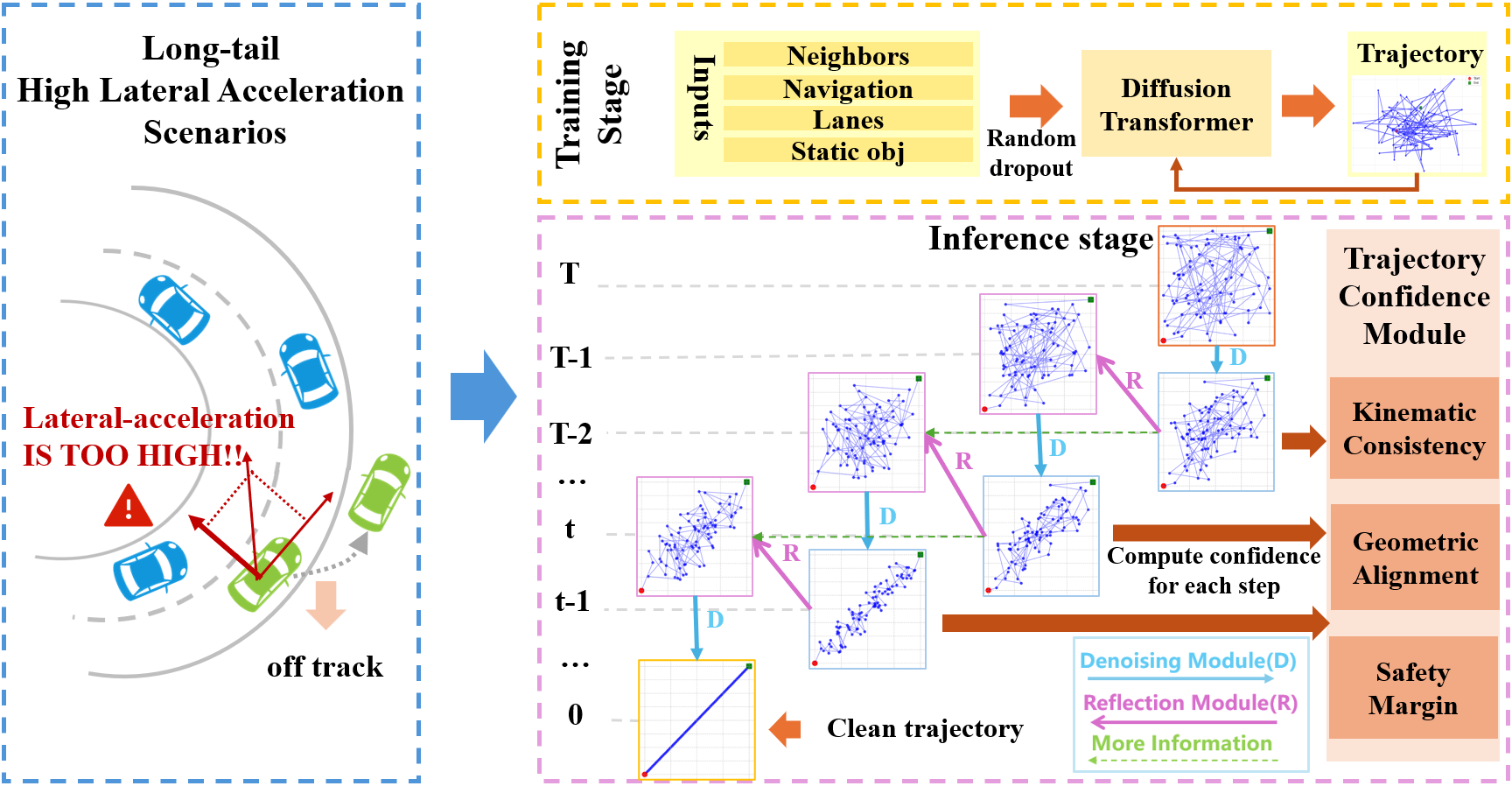
        } 
        \caption{Architecture of ReflexDiffusion. (a) Training Module enhances model robustness to physically-coupled features via a conditional dropout strategy. (b) Denoising Module employs classifier-free guidance to generate initial trajectories. (c) Reflection Module iteratively refines trajectories by injecting physics-aware gradients to enforce curvature-speed-acceleration coupling constraints during inference. (d) Trajectory Confidence Module assesses trajectory reliability, and dynamically triggers the reflection mechanism to ensure safety.}
        \label{fig0}
    \end{figure*}

In this section, we present ReflexDiffusion, which contains four modules to capture and make the most of the underlying characteristics of the long-tail data in training sets, as illustrated in Figure \ref{fig0}.
    \subsection{Training Module}
    \label{sec4.1}
    To address data scarcity in long-tail scenarios, we introduce a conditional dropout strategy \cite{ho_classifier-free_2022} into the Diffusion Planner \cite{zheng_diffusion-based_2025} training framework. While preserving its unified modeling capacity for neighbor vehicles, lane structures, and navigation signals, our approach strategically drops critical physics-coupled features during training to simulate real-world sensory degradation. Crucially, this includes randomly masking road turning radius $R$ and vehicle speed $v$ to simulate scenarios where their coupling $a_y = v^2/R$ is obscured. This forces the model to learn robust representations under partial observability, directly addressing the physics-disalignment challenge in extreme maneuvers.
    Given a training sample with (1) $\mathbf{x}_{gt}$: Ground-truth trajectory; (2) $\mathbf{c}_{full} = [\mathbf{c}_{{neighbors}}, \mathbf{c}_{{lanes}},\mathbf{c}_{{nav}},  \mathbf{c}_{{static\_obj}}]$: Condition vector (neighbor vehicle data, lanes, navigation information, static objects); (3) $\mathbf{c}_{decouple}=[\mathbf{c}_{nav}]$, which drops lane information such as road turning radius $R$ and agent data including ego vehicle speed $v$.

We generate masked condition $\tilde{\mathbf{c}}$ via 10\% conditional dropout:
\begin{equation}
\tilde{\mathbf{c}} = 
\begin{cases} 
\mathbf{c}_{decouple} & \text{with probability } p_{\text{drop}}\\
\mathbf{c}_{full} & \text{otherwise}
\end{cases}
\label{eq1}
\end{equation}
The diffusion training objective then becomes:
\begin{equation}
\mathcal{L}_{\theta} = \mathbb{E}_{t \sim \mathcal{U}(0,T), \mathbf{x}_{t} \sim q_{t_0}(\textbf{x}_t\ |\textbf{x}_{gt}, \tilde{\mathbf{c}})} \left[ \| \mathbf{x}_{gt} - \mathbf{x}_t \|^2 \right]
\label{eq1.5}
\end{equation}
while the diffusion training objective in Diffusion Planner is:
\begin{equation}
  \mathcal{L}_{\theta} = \mathbb{E}_{t \sim \mathcal{U}(0,T), \mathbf{x}_{t} \sim q_{t_0}(\textbf{x}_t\ |\textbf{x}_{gt}, \mathbf{c}_{{full}})} \| \mathbf{x}_{gt} - \mathbf{x}_t) \|^2
  \label{eq3}
\end{equation}
    \subsection{Denoising Module}

To address trajectory quality degradation in long-tail scenarios, particularly the inability to handle non-differentiable safety constraints, existing SOTA Diffusion Planners \cite{zheng_diffusion-based_2025} rely on complex classifier guidance (CG) during denoising. This approach requires handcrafting a guidance function $f_{\text{cg}}(\tau)$ that must be
continuously differentiable, multi-objective and computationally tractable.

However, these requirements lead to oversimplified approximations, severely limiting performance in safety-critical edge cases. Therefore, we utilize classifier-free guidance (CFG) \cite{ho_classifier-free_2022} as a principled alternative. The key innovation lies in embedding trajectory curvature constraint awareness directly into the diffusion model through conditional dropout during training, eliminating the need for auxiliary guidance functions. The CFG denoising process is formalized as:
\begin{equation}
\begin{aligned}
\epsilon_{\theta}^t (\mathbf{x}_t) ={}& \epsilon_{\theta}(\mathbf{x}_t \mid \mathbf{c}_{{decouple}}) \\
&+ \lambda_{1} \left[ \epsilon_{\theta}(\mathbf{x}_t \mid \mathbf{c}_{{full}}) - \epsilon_{\theta}(\mathbf{x}_t \mid \mathbf{c}_{{decouple}}) \right]
\end{aligned}
\label{eq4}
\end{equation}

\begin{equation}
\mathbf{x}_{t-1} = \frac{1}{\sqrt{\alpha_t}} \left( \mathbf{x}_t - \frac{\beta_t}{\sqrt{1-\bar{\alpha}_t}} \epsilon_{\theta}^t(\mathbf{x}_t) + \sigma_t \mathbf{z} \right)
\label{eq5}
\end{equation}
where $\epsilon_{\theta}^t(\mathbf{x}_t)$ represents the noise predicted by the CFG at time t, $\lambda_{1}$ is the guidance scale controlling constraint adherence intensity, $\beta_t=1-\alpha_t$, $\bar{{\alpha}_t}=\alpha_1\alpha_2...\alpha_t$, and $\mathbf{z} \sim \mathcal{N}(0,1)$.

    \subsection{Reflection Module}
Inspired by the reflective reasoning in large language models (LLMs) that iteratively refine outputs via self-critique \cite{ji_towards_2023, madaan_self-refine_2023}, we are the first to propose the reflection mechanism for autonomous driving trajectory generation. This paradigm shift addresses inadequate long-tail information utilization in standard diffusion denoising, where single-pass sampling fails to recover critical scene semantics in rare events. Conventional denoising suffers from (1) Information dilution: Progressive noise removal attenuates long-tail features. (2) Error accumulation: Early-stage deviations amplify in later steps. (3) No recourse: Once generated, low-quality trajectories cannot be revised.

Our physics-aware reflection mechanism breaks this linear flow through iterative information boosting. After obtaining $x_{t-1}$ from standard CFG denoising (Eq.~\ref{eq5}), to make the next reflection deterministic, we use the DDIM scheduler \cite{DBLP:conf/iclr/SongME21} for denoising, for which the denoising formula becomes:
\begin{equation}
\mathbf{x}_{t-1} = \frac{1}{\sqrt{\alpha_t}} \left( \mathbf{x}_t - \frac{\beta_t}{\sqrt{\alpha_t - \bar{\alpha_t}} + \sqrt{1 - \bar{\alpha_t}}} \epsilon_\theta^t(\mathbf{x}_t)  \right)
\label{eq2}
\end{equation}

After obtaining $\mathbf{x}_{t-1}$ from CFG denoising, we compute trajectory confidence $\mathcal{C}(\mathbf{x}_{t-1})$. If $\mathcal{C}(\mathbf{x}_{t-1}) < \gamma$, reflection mechanism is triggered. We approximate the noise predicted at timestep $t$ with timestep $t-1$ along the reflection path, i.e, set $\epsilon_\theta^t(\mathbf{x}_t)\approx \epsilon_\theta^t(\mathbf{x}_{t-1})$.
\begin{equation}
\mathbf{x}'_t=\sqrt{\alpha_t} \mathbf{x}_{t-1}+\frac{\beta_t}{\sqrt{\alpha_t - \bar{\alpha_t}} + \sqrt{1 - \bar{\alpha_t}}} \epsilon_\theta^t\left(\mathbf{x}_{t-1}\right)
\end{equation}
$\epsilon_\theta^t(\mathbf{x}_{t-1})$ contains information about the coupling of road curvature $\kappa$ and vehicle speeds $v$, which can be rewritten as $\Delta_{{couple}}$:
\begin{equation}
\begin{aligned}
\Delta_{{couple}} =& \underbrace{\epsilon_{\theta}(\mathbf{x}_{t-1} \mid \mathbf{c}_{{decouple}})}_{\text{exploration}} \\
+&\lambda_{2} \underbrace{(\epsilon_{\theta}(\mathbf{x}_{t-1} \mid \mathbf{c}_{{full}}) - \epsilon_{\theta}(\mathbf{x}_{t-1} \mid \mathbf{c}_{{decouple}}))}_{\text{condition gradient}} 
\end{aligned}
\end{equation}
where $\lambda_{2}$ is the reflection guidance scale controlling constraint adherence intensity.\\
Projecting $\Delta_{{couple}}$ onto centripetal constrained manifolds:
\begin{equation}
\Delta_{{proj}} = \mathbf{P} \cdot \Delta_{{couple}}, \quad \mathbf{P} = \begin{bmatrix} 
\frac{\partial (\kappa v^2)}{\partial \kappa} & \frac{\partial (\kappa v^2)}{\partial v} \\ 
0 & 1 
\end{bmatrix}
\end{equation}
The projection matrix $\mathbf{P}$ enforces physical constraints by mapping gradient adjustments $\Delta_{{couple }}$ onto the centripetal force manifold $a_y \approx \kappa v^2$. The projection matrix P has the structure:
\begin{itemize}
\item Top row $\left(v^2, 2 \kappa v\right)$ : Amplifies curvature $(\kappa)$-velocity $(v)$ coupling using partial derivatives of $\kappa v^2$.
\item Bottom row $(0,1)$ : Preserves unconstrained motion freedom. This ensures trajectory corrections strictly satisfy vehicle dynamics while maintaining distributional coherence through iterative denoising:
\begin{equation}
\mathbf{x}'_t = \sqrt{\alpha_t}\mathbf{x}_{t-1} + b \cdot \Delta_{\text{proj}}
\label{eq8}
\end{equation}
\end{itemize}
The reflection sample $\mathbf{x}'_t$ adds more information about road curvature $\kappa$ and vehicle speeds $v$ compared to the pre-reflection sample $\mathbf{x}_t$:
\begin{equation}
\begin{aligned}
\mathbf{x}_t^{\prime} =\mathbf{x}_t-\frac{\beta_t}{\sqrt{\alpha_t - \bar{\alpha_t}} + \sqrt{1 - \bar{\alpha_t}}} \epsilon_\theta^t\left(\mathbf{x}_t\right)+b \cdot \Delta_{\mathrm{proj}}
\end{aligned}
\end{equation}
\begin{table*}
        [htbp]
        \centering
        \begin{tabular}{c|c|cccc|cccc}
            \hline
            \multirow{2}{*}{Type}           & \multirow{2}{*}{Planner}  &
            & \multicolumn{2}{c}{Test14-Hard} & &  & \multicolumn{2}{c}{Test14-Random} & \\
            \cline{3-10}  & & & NR & R & & & NR  & R &  \\
            \hline
            \multirow{2}{*}{Rule-based}     
            & IDM & & 36.74  & 62.42 & & & 67.61 & 64.66 \\
            & PDM-Closed & & 32.55 & 53.03 & & & 75.69 & 82.17\\
            \hline
            \multirow{3}{*}{Hybrid}
            & PDM-Hybrid & & 32.54 & 53.04 & & & 75.97 & 82.17 \\
            & Gameformer & & 53.12 & 57.46 & & & 82.60 & 79.49 \\
            & SAH-Drive & & 43.08 & 57.40 & & & \textbf{91.18} & \textbf{89.27} \\
            \hline
            \multirow{9}{*}{Learning-based} 
            & UrbanDriver & & 26.09   & 14.66 & & & 29.17  & 33.27  \\
            & Diffusion-es & & 44.63   & 52.72 & & & 88.20  & 84.20  \\
            & PlanCNN & & 20.16 & 28.03 & & & 48.20 & 51.38 \\
            & PlanTF       & & 38.24  & 42.52 & & & 73.66  & 67.81 \\
            & Pluto        & & 42.21  & 45.98 & & & 81.67  & 75.95  \\
            & Diffusion Planner & & 58.47 & 57.41 & & & 71.60 & 82.88 \\
            & Diffusion Planner w/ conditional dropout & & 12.60 & 14.04 & & & 38.45 & 44.88 \\
            & Diffusion Planner w/ conditional dropout+cfg & & 44.40 & 55.55 & & & 76.44 & 70.21 \\
            & \textbf{Ours}  & & \textbf{59.94}  & \textbf{65.53} & & & 86.40 & 71.57 \\
            \hline
        \end{tabular}
        \caption{Closed-loop planning results on high-lateral-acceleration scenarios in the long-tail dataset and regular dataset. To ensure that the performance gains are solely attributed to our proposed reflection mechanism and not training strategies, we retrain the current SOTA baseline Diffusion Planner, under two additional settings: Conditional dropout and Classifier-free guidance (CFG) for a comprehensive and fair comparison. Our method achieves SOTA on Test14-hard high-lateral-acceleration scenarios. NR: non-reactive mode. R: reactive mode.}
        \label{table1}
    \end{table*}
Compared to prior works, our reflection mechanism provides distinct advantages. Unlike classifier guidance (CG) or classifier-free guidance (CFG) methods that rely on one-shot generation, limiting their capacity for error correction, our approach enables iterative refinement through cyclic noise reversion. Whereas RL-based planners require costly retraining to adapt to new scenarios, our solution operates purely during inference without parameter updates. While ensemble methods achieve robustness via multiple models at the expense of high computational overhead, our method maintains efficiency by utilizing a single model architecture with minimal latency increase.
 
    \subsection{Trajectory Confidence Module}

To assess the reliability of generated trajectories throughout the denoising and reflection cycles, we introduce a multi-factor confidence metric. This module fulfills two key roles: (1) determining when to trigger reflection, and (2) ensuring safety through real-time risk evaluation to enable effective fallback mechanisms.

The confidence metric $\mathcal{C}(\mathbf{x}_t)$ quantifies the reliability of generated trajectories in high-lateral-acceleration scenarios, integrating three domain-specific factors:

\label{eq:confidence}

\subsubsection*{1. Kinematic Consistency ($\mathcal{D}_{\text{kin}}$).}
Assesses compliance with fundamental motion constraints:
\begin{equation}
\begin{aligned}
\mathcal{D}_{\text{kin}} = &\exp\left( -m_1 |a_y^{\text{traj}} - a_y^{\text{ref}}| \right) \cdot \sigma\left( j_{\max} - |j_{\text{lat}}| \right)\\
& a_y^{\text{traj}}= \kappa_{\tau} v^2, j_{\text{lat}} = \frac{da_y}{dt}
\end{aligned}
\end{equation}

\subsubsection*{2. Geometric Alignment ($\mathcal{G}_{\text{align}}$).}
Quantifies adherence to road geometry:
\begin{equation}
\begin{aligned}
\mathcal{G}_{\text{align}} &= \mathcal{R}_{\text{curv}} \times \mathcal{R}_{\text{dev}} \\
\mathcal{R}_{\text{curv}} &= \exp\left( -m_2 |\kappa_{\tau} - \kappa_{\text{road}}| \cdot R_{\text{curve}} \right) \\
\mathcal{R}_{\text{dev}} &= 1 - \frac{\max(0, d_{\text{dev}} - d_{\text{safe}})}{d_{\max}} \\
\end{aligned}
\end{equation}
where $R_{\text{curve}} = 1/|\kappa_{\text{road}}|$: Current road curve radius, $d_{\text{dev}} = \max \| \mathbf{p}_{\tau} - \mathbf{p}_{\text{center}} \|_2$: Maximum lateral deviation.

\subsubsection*{3. Safety Margin ($\mathcal{S}_{\text{margin}}$).}
Evaluates risk buffers:
\begin{equation}
\mathcal{S}_{\text{margin}} = \sigma\left( \frac{\text{TTC} - 2.5}{0.5} \right) \cdot (1 - p_{\text{ODA}}) \cdot \cos(\Delta\psi)
\end{equation}
where TTC represents time-to-collision with nearest obstacle, $p_{\text{ODA}}$ is probability of out-of-drivable-area violation, and $\Delta\psi$ is heading deviation from road tangent.

    \section{Experiments}
    In this section, we conduct extensive experiments to demonstrate the performance
    of our method.
    \subsection{Experimental Setup}
    \textbf{Datasets.}
    To assess the performance of ReflexDiffusion on high-lateral-acceleration scenarios in long-tail dataset, we utilize Test14-hard \cite{cheng_rethinking_2024} and Test14-random \cite{cheng_rethinking_2024} as our benchmarks. Both benchmarks are developed to cover the 14 scenario types including our target scenarios in the nuPlan Challenge, with the former introduced especially to represent long-tail scenarios.\\
    \textbf{Baselines.}
    The baselines are mainly divided into three groups \cite{dauner_parting_nodate}: Rule-based \cite{treiber_congested_nodate, dauner_parting_nodate}, Learning-based \cite{urbandrive, yang_diffusion-es_nodate, plancnn, cheng_rethinking_2024, cheng_pluto_2024, zheng_diffusion-based_2025}, and Hybrid methods \cite{dauner_parting_nodate, gameformer, sah}, which incorporate additional refinement to the outputs of the learning-based model. We compare ReflexDiffusion against
    the baselines above, which are detailed in Appendix.
    \begin{figure}[htbp]
        \centering
        \subfigure[Diffusion Planner]{
        \begin{minipage}[t]{0.5\linewidth}
        \centering
        \includegraphics[width=1.5in]{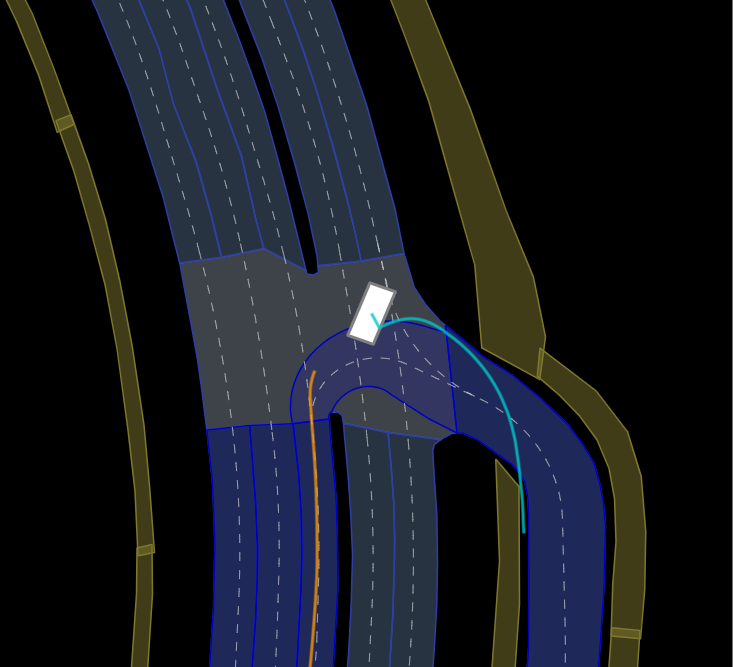}
        \end{minipage}%
        }%
        \subfigure[ReflexDiffusion]{
        \begin{minipage}[t]{0.5\linewidth}
        \centering
        \includegraphics[width=1.5in]{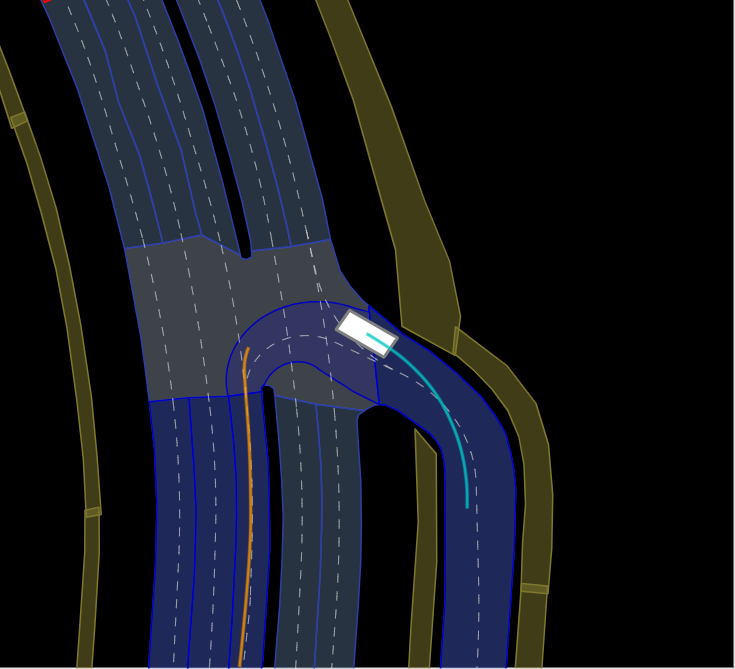}
        \end{minipage}%
        }%
        \centering
        \caption{Visualization comparison in U-turn scenario from high-lateral-acceleration scenarios. The Diffusion Planner's trajectory veers out of the lane during the turn, while ReflexDiffusion makes the turn without incident!}
        \label{fig1}
        \end{figure}
    \subsection{Main Results}
\begin{figure}[t]
        \centering
        \includegraphics[width=0.99\columnwidth]{
            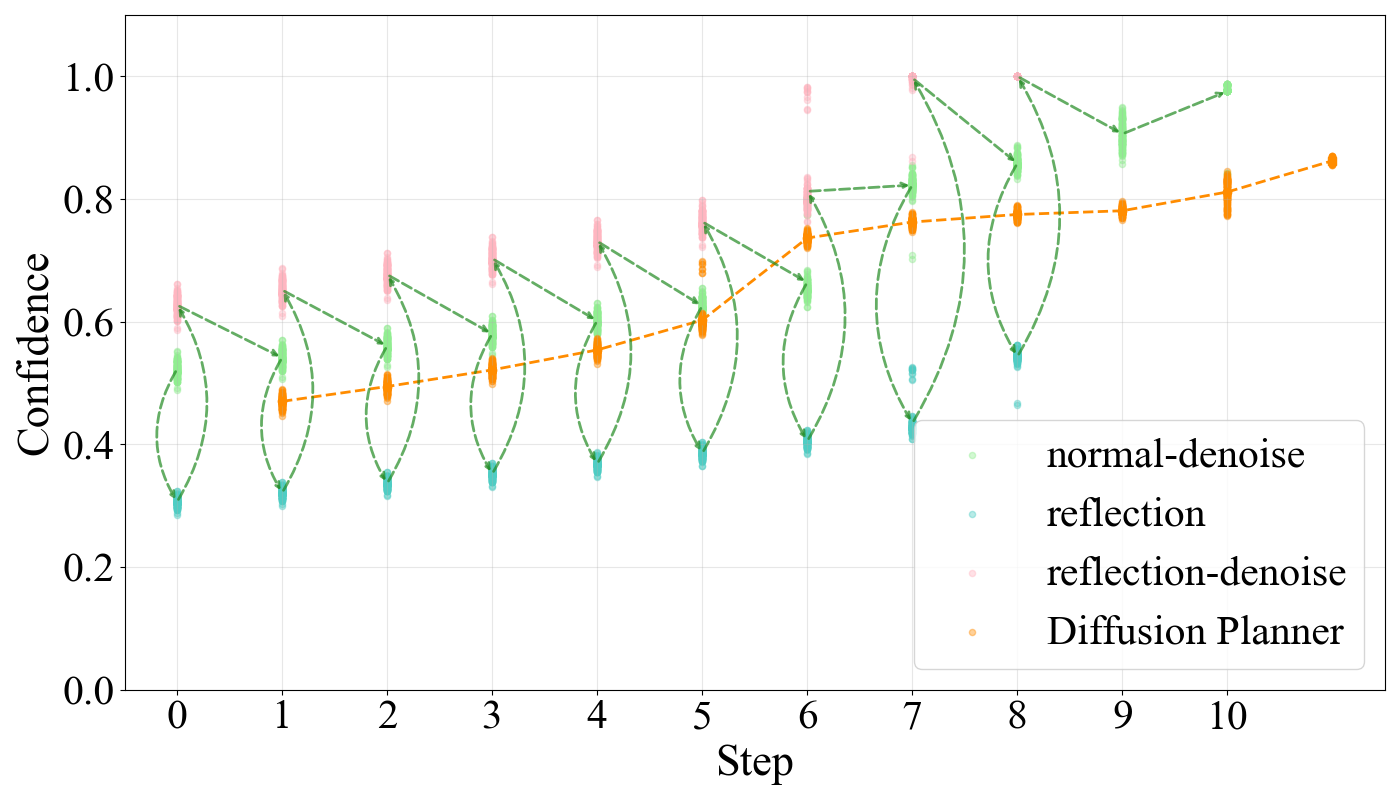
        } 
        \caption{Visualization of trajectory confidence in inference process. 
        \tikz \draw[color=green!70!black, dashed, line width=1pt, dash pattern=on 3pt off 2pt] (0,0.5ex) -- (0.5cm,0.5ex); represents the trend of trajectory confidence for ReflexDiffusion and \tikz \draw[color=orange, dashed, line width=1pt, dash pattern=on 3pt off 2pt] (0,0.5ex) -- (0.5cm,0.5ex); represents the trend of trajectory confidence for Diffusion Planner.}
        \label{fig4.5}
        \end{figure}

    We compare the performance of ReflexDiffusion with baselines on high-lateral-acceleration scenarios in the long-tail dataset Test14-hard, as well as the regular dataset Test14-random.
    The quantitative and qualitative results are shown in Table \ref{table1} to Table \ref{table4} and Figure \ref{fig1} to Figure \ref{fig4.5}, from which we derive the following findings.
    
    \textbf{Our method achieves state-of-the-art (SOTA) performance on high-lateral-acceleration scenarios in long-tail benchmarks while maintaining strong performance on regular datasets.} As summarized in Table \ref{table1}, ReflexDiffusion outperforms the current SOTA Diffusion Planner on the Test14-hard benchmark in both non-reactive and reactive modes, achieving a significant 14.1\% improvement in driving score for the reactive mode. Concurrently, performance on the regular dataset is also substantially enhanced, with a 20.7\% improvement in the non-reactive mode. Runtime analysis conducted on a single RTX 4090 GPU (Table \ref{table1.3}) indicates that enabling the reflection mechanism increases the per-step latency from 3.3 ms to 6.3 ms. However, since reflection is triggered in only $\leq$ 0.5\% of evaluated real-world driving cases, the average runtime is calculated as $0.5\% \times 122.7 + 99.5\% \times 35.7 \approx 36.1$ ms. This remains comparable to the SOTA baseline, sustaining a control frequency of $>$ 20 Hz suitable for real-time deployment. Crucially, this marginal increase in latency is traded for a substantial gain in planning robustness and safety.
       
    \begin{table}[t]
       \centering
       \begin{tabular}{ccc}
            \hline
            Planner & per-step & e2e \\ 
            \hline
            Diffusion-es & / & 7612.7  \\
            Diffusion Planner   & / & 35.7                \\
            ReflexDiffusion w/o reflection & 3.3 & 35.7 \\
            \textbf{Ours} & \textbf{6.3}    & \textbf{36.1} \\                                                                   
            \hline
        \end{tabular}
       \caption{Runtime tests among diffusion-based planners. The runtime of ReflexDiffusion stays near the current SOTA baseline, and trades minor latency for higher robustness.}
       \label{table1.3}
       \end{table}
       
         \textbf{The effectiveness of our method is further demonstrated through trajectory confidence.} 
        The superior performance of ReflexDiffusion in high-lateral-acceleration scenarios is corroborated by trajectory confidence analysis. In a representative U-turn scenario where the baseline Diffusion Planner fails, our method achieves a significantly higher driving score (Table \ref{table1.5}). A qualitative comparison in Figure \ref{fig1} illustrates this improvement: while the baseline planner generates trajectories that violate drivable areas due to excessive lateral acceleration, ReflexDiffusion produces trajectories that maintain both safety and feasibility.
       
        We attribute the baseline's failures to an insufficient representation of extreme vehicle dynamics in the training data, which critically impairs generalization to high-lateral-acceleration cases. To address this, ReflexDiffusion introduces a reflective adjustment step during inference. After each denoising iteration, a physics-aware gradient—derived from the difference between conditional and unconditional predictions—is computed and injected to explicitly amplify critical physical constraints. Through this iterative refinement, trajectories progressively converge to satisfy the centripetal force constraint $a_y \leq v^{2}/R$. As visualized in the confidence plot (Figure \ref{fig4.5}), the trajectory confidence temporarily decreases during the reflection phase due to exploratory gradient adjustments. This dip signifies an active optimization process against physical constraints. The confidence subsequently recovers and rises to a peak level after the final denoising step. This pattern demonstrates how the reflection mechanism compensates for training data scarcity by dynamically reinforcing vehicle-road interaction physics near handling limits, effectively converting transient uncertainty into enhanced safety margins.

\begin{table}[t]
       \centering
       \begin{tabular}{ccccc}
       \hline
       \multirow{2}{*}{Planner} & \multirow{2}{*}{Score} &\multicolumn{3}{c}{Trajectory Confidence} \\ \cline{3-5} 
       & & nor & re & re-de \\ \hline
       Diffusion Planner & 0.00 & 0.48 & / & /              \\ 
       \textbf{Ours}  & \textbf{100.00} & \textbf{0.87} & \textbf{0.36} & \textbf{0.73}  \\ \hline
       \end{tabular}
       \caption{Driving score comparisons on  U-turn scenario from high-lateral-acceleration scenarios. Including average trajectory confidence of normal-denoising (nor), reflection (re) and reflection-denoising (re-de) process.}
       \label{table1.5}
       \end{table}
       
        \begin{figure*}[t]
        \centering
        \subfigure[Conditional dropout rate]{
        \begin{minipage}[t]{0.24\linewidth}
        \centering
        \includegraphics[width=0.985\textwidth]{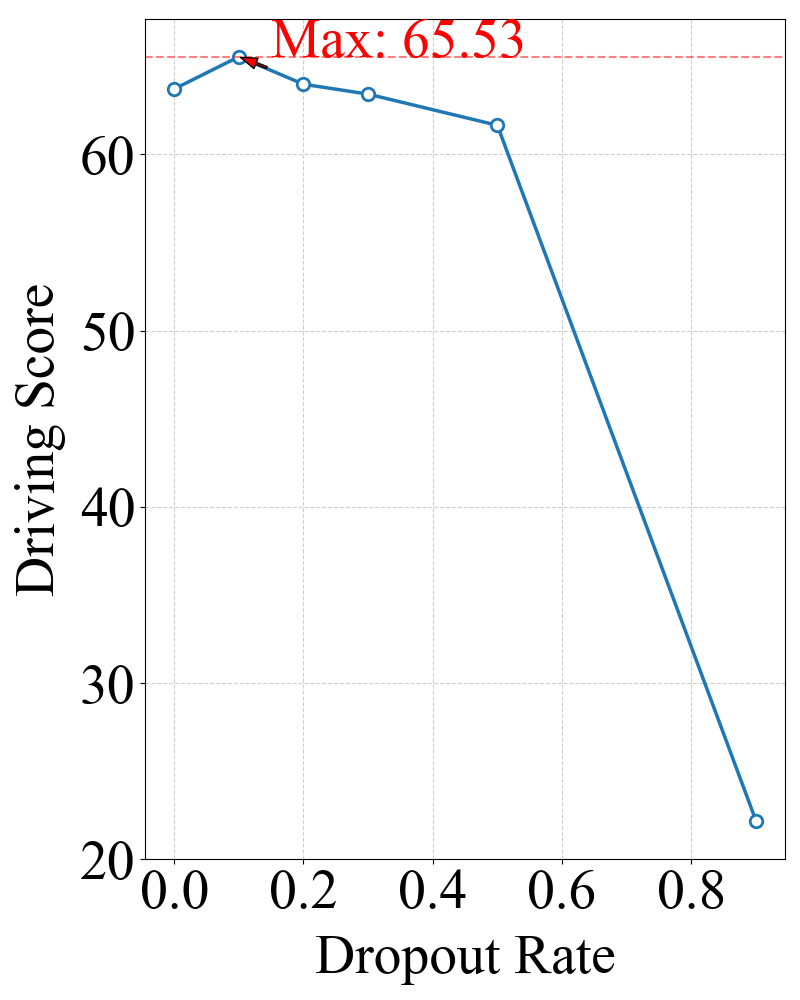}
        \end{minipage}%
        }%
        \subfigure[Denoising scale $\lambda_{1}$]{
        \begin{minipage}[t]{0.24\linewidth}
        \centering
        \includegraphics[width=1\textwidth]{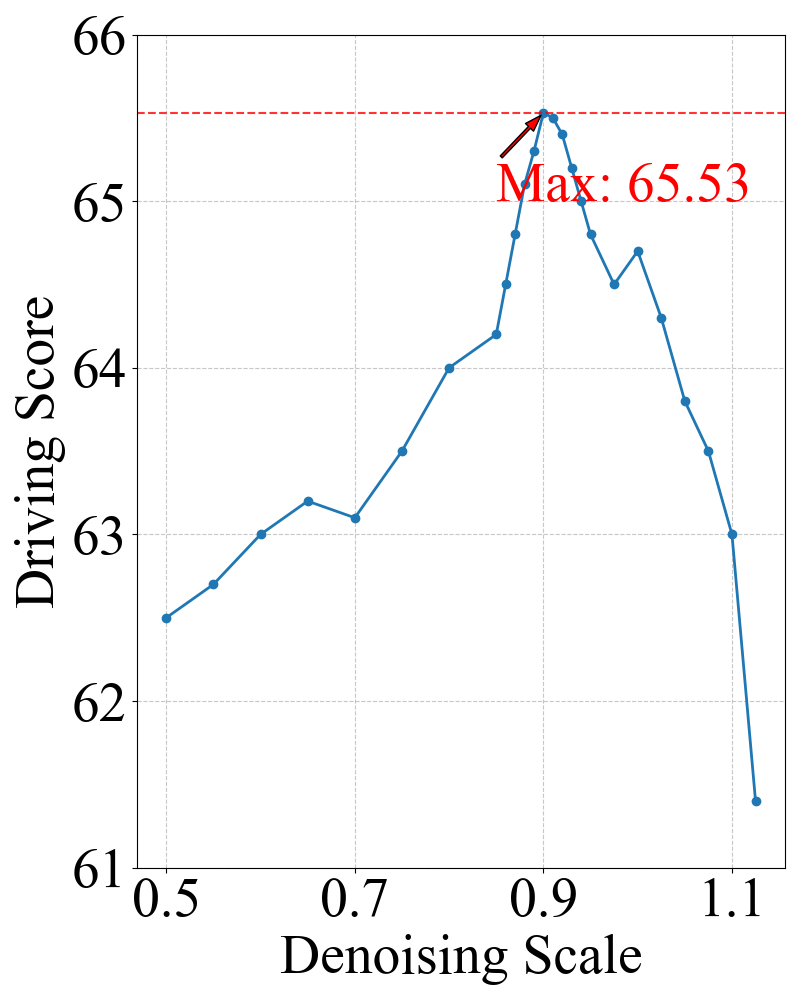}
        \end{minipage}%
        }%
        \subfigure[Reflection scale $\lambda_{2}$]{
        \begin{minipage}[t]{0.24\linewidth}
        \centering
        \includegraphics[width=\textwidth]{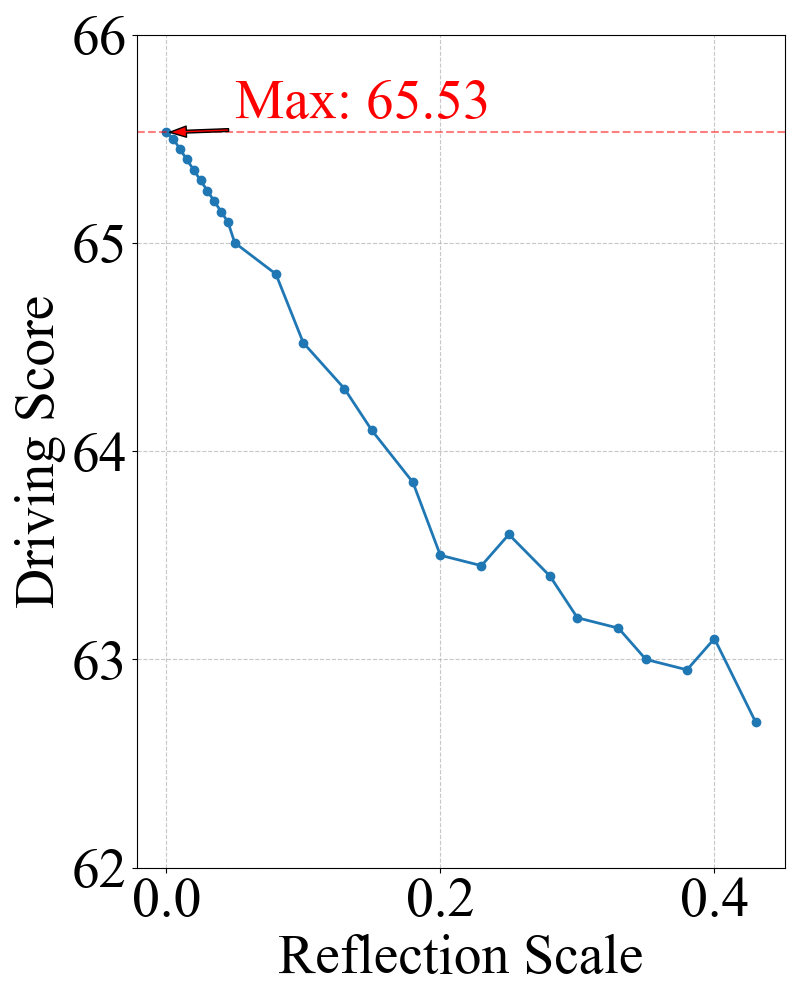}
        \end{minipage}%
        }%
        \subfigure[Confidence threhold $\gamma$]{
        \begin{minipage}[t]{0.24\linewidth}
        \centering
        \includegraphics[width=0.985\textwidth]{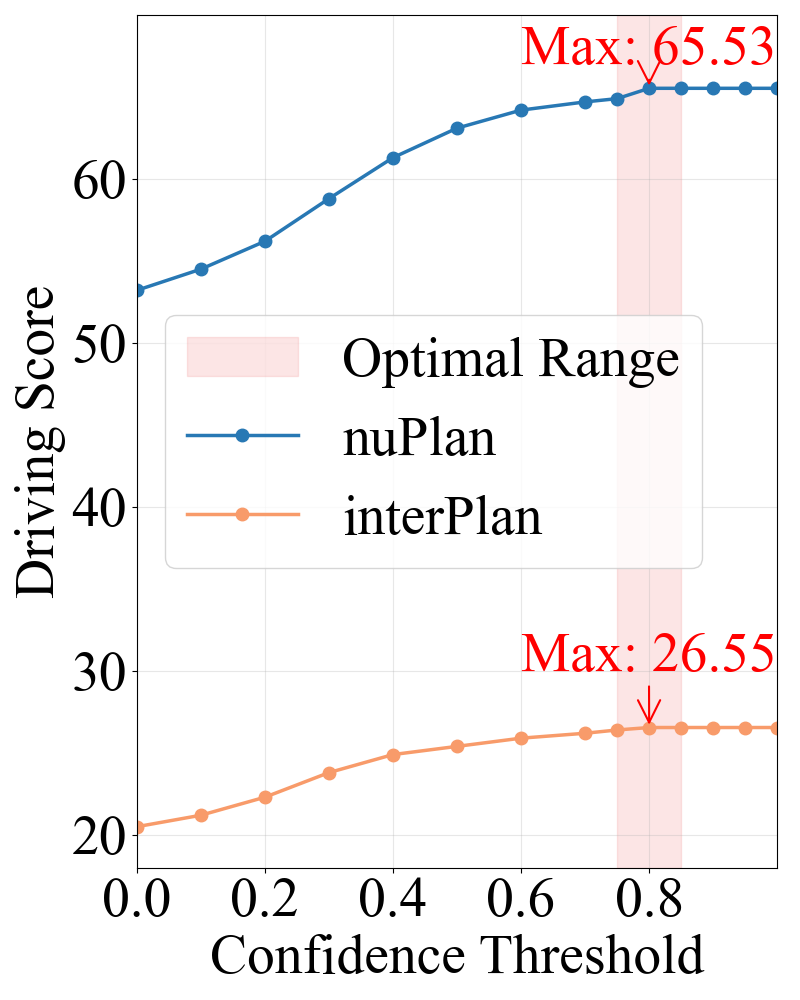}
        \end{minipage}%
        }%
        \centering
        \caption{Ablation Studies. The optimal values for the four parameters: conditional dropout rate, denoising scale $\lambda_1$, reflection scale $\lambda_2$ and confidence threshold $\gamma$ are 0.1, 0.9, 0.0, and 0.8, respectively.}
        \label{fig5}
        \end{figure*}
         \textbf{Our framework demonstrates strong generalizability across different diffusion-based planners.} Our framework is architecture-agnostic, requiring no structural changes to the base planner. Its core innovation lies in an inference-stage reflection mechanism, with the only training modification being a random dropout of input features. This design enables seamless transfer across different diffusion-based planners. Generalization tests on both Diffusion Planner \cite{zheng_diffusion-based_2025} and Diffusion-es \cite{yang_diffusion-es_nodate} confirm its effectiveness, demonstrating significant performance improvements in long-tail scenarios as shown in Table \ref{table2} and Table \ref{table3}. We train the Diffusion-es conditional diffusion model for \textbf{50 epochs} as a baseline, and use the same settings to train a new model with a conditional dropout rate of 10\%. These results validate ReflexDiffusion as a general and practical strategy for enhancing the safety and robustness of existing trajectory planners.
        \begin{table}
        [t]
        \centering
        \begin{tabular}{ccc}
            \hline
            Planner                                    & \begin{tabular}[c]{@{}c@{}}Test14-hard \\ driving score\end{tabular} & \begin{tabular}[c]{@{}c@{}}Test14-random \\ driving score\end{tabular} \\ \hline
            Diffusion Planner   & 57.41 & 71.60                \\
            \textbf{w/ ReflexDiffusion} & \textbf{65.53}                                                           & \textbf{86.40}                                                                         \\
            \hline
        \end{tabular}
        \caption{Generalizability test on Diffusion Planner. The result proves that ReflexDiffusion can significantly enhance the performance of Diffusion Planner in high-risk scenarios.}
        \label{table2}
    \end{table}
    \begin{table}
        [t]
        \centering
        \begin{tabular}{ccc}
            \hline
            Planner                               & \begin{tabular}[c]{@{}c@{}}Test14-hard \\ driving score\end{tabular} & \begin{tabular}[c]{@{}c@{}}Test14-random \\ driving score\end{tabular}  \\
            \hline
            Diffusion-es       & 31.88     & /    \\
            \textbf{w/ ReflexDiffusion} & \textbf{39.04} & /                                                                         \\
            \hline
        \end{tabular}
        \caption{Generalizability test on Diffusion-es. The result indicates that the enhancement effect of ReflexDiffusion is universal and it is a general reasoning stage optimization scheme independent of the model architecture.}
        \label{table3}
    \end{table}

    \subsection{Ablation Studies}
    \textbf{Design Choices for training.} We demonstrate the effectiveness of conditional dropout in the training stage, as shown in Table \ref{table4}, and its optimal dropout rate is demonstrated in Figure \ref{fig5}(a). The results show that the conditional dropout is significant and the optimal performance is achieved at 10\%. \\
    \textbf{Design Choices for inference.} We ablate the classifier-free guidance denoising and reflection mechanism, as shown in Table \ref{table4}, as well as two hyperparameters: denoising scale $\lambda_1$ and reflection scale $\lambda_2$ shown in Figure \ref{fig5}(b) and \ref{fig5}(c), respectively. The results indicate that both classifier-free guidance and reflection mechanisms play a prominent role in enhancing the performance on long-tail datasets, with the optimal combination ($\lambda_1$,  $\lambda_2$)=(0.9, 0). Moreover, the confidence threshold $\gamma$ is empirically calibrated via cross-dataset nuPlan and interPlan \cite{interplan} validation, as shown in Figure \ref{fig5}(d), giving a sensitivity range [0.75,0.85], balancing robustness and responsiveness.
        
    \begin{table}[t]
    \centering
    \begin{tabular}{ccc}
    \hline
    Type                            & Planner             & Score \\ \hline
    Base                            & ReflexDiffusion     & 65.53 \\ \hline
    \multirow{3}{*}{Simplification} & w/o conditional dropout & 23.86 \\  
                                & w/o cfg denoising    & 59.85 \\
                                & w/o reflection & 53.21 \\
                                \hline
    \end{tabular}
    \caption{Ablation of each module during the training and inference process. The result clearly demonstrates that each module is crucial for ReflexDiffusion.}
    \label{table4}
    \end{table}
    
    \section{Conclusion}

This paper presents ReflexDiffusion, a novel framework that pioneers inference-stage reflection for diffusion-based trajectory planning. Inspired by the generate-evaluate-refine
paradigm in large language models, we develop physics-
aware reflection through conditional gradient injection dur-
ing diffusion sampling. This innovation explicitly amplifies
critical physical couplings, particularly the curvature-speed-
acceleration relationship, enabling real-time trajectory cor-
rection without classifier guidance. Comprehensive nuPlan
evaluations demonstrate significant advancements.



\section*{Acknowledgments}
This work is supported by the National Natural Science Foundation of China (NSFC) under grants Nos. 72231011, 72471238, 92370124, 92248303, 62276149.

\bibliography{aaai2026}

\end{document}